\RequirePackage{amsmath}
\documentclass[runningheads]{llncs}
\usepackage[utf8]{inputenc}

\usepackage{graphicx}
\usepackage{amssymb, amsthm, mathtools}
\usepackage{physics}
\usepackage{array, multirow}
\usepackage{calc}
\usepackage{tabularx}
\usepackage[hmarginratio=1:1]{geometry}
\usepackage{colortbl}
\usepackage{tikz}
	\usetikzlibrary{positioning}
	\usetikzlibrary{arrows.meta}
\usepackage{xspace}
\usepackage{pgfplotstable}
\usepackage{booktabs}
\usepackage{enumitem}
\usepackage{numprint}

\newcommand{\Omit}[1]{}
\newcommand{\yt}{\textit{YouTube-8M}\xspace}
\newcommand{\F}{{\ensuremath \mathcal F}}

\newcolumntype{L}[1]{>{\raggedright\arraybackslash}p{#1}}
\newcolumntype{C}[1]{>{\centering\arraybackslash}p{#1}}

\begin{document}
\title{Training compact deep learning models for video classification using circulant matrices}

%
%
\author{Alexandre Araujo\inst{1, 2} \and
  Benjamin Negrevergne\inst{1} \and
  Yann Chevaleyre\inst{1} \and
  Jamal Atif\inst{1}}
\authorrunning{}
%
\institute{PSL, Université Paris-Dauphine, LAMSADE, CNRS, UMR 7243, Paris, France
\and
Wavestone, Paris, France}
\maketitle              
\begin{abstract}

In real world scenarios, model accuracy is hardly the only factor to consider. Large models  consume more memory and are computationally more intensive, which makes them difficult to train and to deploy, especially on mobile devices. In this paper, we build on recent results at the crossroads of Linear Algebra and Deep Learning which demonstrate how imposing a structure on large weight matrices can be used to reduce the size of the model. We propose very compact models for video classification based on state-of-the-art network architectures such as \textit{Deep Bag-of-Frames}, \textit{NetVLAD} and \textit{NetFisherVectors}. We then conduct thorough experiments using the large \yt video classification dataset. As we will show, the  circulant DBoF embedding achieves an excellent trade-off between size and accuracy.

\end{abstract}

\section{Introduction}
\label{sec:intro}

The top-3 most accurate approaches proposed during the first \yt\footnote{https://www.kaggle.com/c/youtube8m} video classification challenge  were all ensembles models. The ensembles typically combined models based on a variety of deep learning architectures such as \textit{NetVLAD}, \textit{Deep Bag-of-Frames} (DBoF), \textit{NetFisherVectors} (NetFV) and \textit{Long-Short Term Memory} (LSTM), leading to a large aggregation of models (25 distinct models have been used by the first contestant~\cite{DBLP:journals/corr/MiechLS17}, 74 by the second~\cite{DBLP:journals/corr/WangZW17} and 57 by the third~\cite{DBLP:journals/corr/LiGLBLLLZW17}). Ensembles are accurate, but they are not ideal: their size makes them difficult to maintain and deploy, especially on mobile devices. 

A common approach to compress large models into smaller ones is to use {\em model distillation}~\cite{44873}. Model distillation is a two steps training procedure: first, a large model (or an ensemble model) is trained to be as accurate as possible. Then, a second compact model is trained to approximate the first one, while satisfying the given size constraints. The success of model distillation and other model compression techniques begs an important question: is it possible to devise models that are compact by nature while exhibiting the same generalization properties as large ones?

In linear algebra, it is common to exploit structural properties of matrices to reduce the memory footprint of an algorithm. 
Cheng et al.~\cite{7410684} have used this principle in the context of deep neural networks to design compact network architectures by imposing a structure on weight matrices of fully connected layers. They were able to replace large, unstructured weight matrices with structured \textit{circulant matrices} without significantly impacting the accuracy. And because a n-by-n circulant matrix is fully determined by a vector of dimension $n$, they were able to train a neural network using only a fraction of the memory required to train the original network.

Inspired by this result, we designed several compact neural network architectures for video classification based on standard video architectures such as NetVLAD, DBoF, NetFV and we evaluated them on the large \yt dataset. However, instead of adopting the structure used by \cite{7410684} (initially proposed by \cite{VYBIRAL20111096}), we decomposed weight matrices into products of diagonal and circulant matrices (as in \cite{schmid2000decomposing}). In contrast with \cite{VYBIRAL20111096} which has been proved to approximate distance preserving projections, this structure can approximate \textit{any} transformation (at the cost of a larger number of weights). As we will show, this approach exhibits good results on the video classification task at hand. 

In this paper, we bring the following contributions:

\begin{itemize}[noitemsep, topsep=0pt]
  \item We define a compact architecture for video classification based on circulant matrices. As a side contribution, we also propose a new pooling technique which improves the Deep Bag-of-Frames embedding. 
  \item We conduct thorough experimentations to identify the layers that are less impacted by the use of circulant matrices and we fine-tune our architectures to achieve the best trade-off between size and accuracy.  
  \item We combine several architectures into a single model to achieve a new model trained-end-to-end that can benefit from architectural diversity (as in ensembles).
  \item We train all our models on the Youtube-8M dataset with the 1GB model size constraint imposed by the \textit{2nd YouTube-8M Video Understanding Challenge}\footnote{https://www.kaggle.com/c/youtube8m-2018}, and compare the different models in terms of size vs. accuracy ratio. Our experiments demonstrate that the best trade-off between size and accuracy is obtained using circulant DBoF embedding layer.

\end{itemize}

\section{Related Works}
\label{sec:related-work}

Classification of unlabeled videos streams is one of the challenging tasks for machine learning algorithms.  Research in this field has been stimulated by the recent release of several large annotated video datasets such as \textit{Sports-1M}~\cite{karpathy2014large}, \textit{FCVID}~\cite{FCVID} or the \yt~\cite{45619} dataset.

The naive approach to achieve video classification is to perform frame-by-frame image recognition, and to average the results before the classification step. However, it has been shown in \cite{45619,DBLP:journals/corr/MiechLS17} that  better results can be obtained by building features across different frames and several deep learning architectures have been designed to learn embeddings for sets of frames (and not single frames). For example Deep Bag-of-Frames ~(DBoF)~\cite{45619}, NetVLAD~\cite{Arandjelovic16} or architectures based on Fisher Vectors~\cite{4270291}.

The DBoF embedding layer, proposed in~\cite{45619} processes videos in two steps. 
First, a learned transformation projects all the frames together into a high dimensional space. 
Then, a max (or average) pooling operation aggregates all the embedded frames into a single discriminative vector  representation of the video.  The NetVLAD~\cite{Arandjelovic16} embedding layer is built on {\em VLAD}~\cite{jegou:inria-00548637}, a technique that aggregates a large number of local frame descriptors into a compact representation using a codebook of visual words. In NetVlad, the codebook is directly learned end-to-end during training. Finally, NetFisherVector (NetFV) is inspired by~\cite{4270291} and uses  first and second-order statistics as video descriptors also gathered in a codebook. The technique can benefit from deep learning by using a deep neural network to learn the codebook \cite{DBLP:journals/corr/MiechLS17}.

All the architectures mentioned above can be used to build video features in the sense of features that span across several frames, but they are not designed to exploit the sequential nature of videos and capture motion.  In order to learn truly spatio-temporal features and account for motion in videos, several researchers have looked into recurrent neural networks (e.g. LSTM~\cite{yue2015beyond,DBLP:journals/corr/LiGLBLLLZW17}) and 3D convolutions~\cite{karpathy2014large} (in space and time).
However, these approaches do not  outperform non-sequential models, and  the single best model proposed in~\cite{DBLP:journals/corr/MiechLS17} (winner of the first \yt competition) is based on NetVLAD~\cite{Arandjelovic16}. 

\paragraph{}
The \textit{2nd YouTube-8M Video Understanding Challenge} includes a constraint on the model size and many competitors have been looking into building efficient memory models with high accuracy.
There are two kinds of techniques to reduce the memory required for training and/or inference in neural networks. The first kind aims at \textit{compressing} an existing neural network into a smaller one, (thus it only impacts the size of the model at inference time). The second one aims at {\em constructing models that are compact} by design. 

To compress an existing network several researchers have investigated techniques to prune parameters that are redundant (e.g. ~\cite{pmlr-v80-dai18d,han2015deep_compression,NIPS2017_6813}). Redundant parameters can be omitted from the model without significantly changing the accuracy. It is also possible to use sparsity regularizers during training, to be able to compress the model after the training using efficient sparse matrix representations (e.g.~\cite{Collins2014MemoryBD,pmlr-v80-dai18d,7298681}). Building on the observation that weight matrices are often redundant, another line of research has proposed to use matrix factorization~\cite{NIPS2013_5025,Jaderberg2014SpeedingUC,8099498} in order to decompose large weight matrices into factors of smaller matrices before inference. 

An important idea in model compression, proposed by Buciluǎ et al.~(\cite{bucilua2006model}), is based on the observation that the model used for training is not required to be the same as the one used for inference. First, a large complex model is trained using all the available data and resources  to be as accurate as possible, then a smaller and more compact model is trained to approximate the first model. The technique which was later specialized for deep learning models by~\cite{44873} (a.k.a. model distillation) is often used to compress large ensemble models into compact single deep learning models.

Instead of compressing the model after the training step, one can try to design models that are compact by nature (without compromising the generalization properties of the network). The benefit of this approach is that it reduces  memory usage required during both training and inference. As a consequence, users can train models that are virtually larger using less time and less computing resources. They also save the trouble of training two models instead of one as it is done with distillation. These techniques generally work by constraining the weight representation, either at the level of individual weights (e.g. using floating variable with limited precision \cite{Gupta:2015:DLL:3045118.3045303}, quantization \cite{Courbariaux:2015:BTD:2969442.2969588,DBLP:journals/corr/MellempudiKM0KD17,rastegariECCV16}) or  at the level of the whole matrix, (e.g. using weight hashing techniques~\cite{Chen:2015:CNN:3045118.3045361}) which can achieve better compression ratio. However in practice, hashing techniques are difficult to use because of their irregular memory access patterns which makes them inadequate for GPU-execution.

Another way of constraining the weight representation is to impose a structure on weight matrices (e.g. using circulant matrices~\cite{7410684,NIPS2015_5869}, Vandermonde~\cite{NIPS2015_5869} or Fastfood transforms~\cite{7410530}). In this domain, Cheng et al. ~\cite{7410684} have proposed to replace two fully connected layers of AlexNet by circulant and diagonal matrices where the circulant matrix is learned by a gradient based optimization algorithm and the diagonal matrix entries are sampled at random in \{-1, 1\}. The size of the model is reduced by a factor of 10 without loss in accuracy\footnote{In network such as AlexNet, the last 3 fully connected layers use 58M out of the 62M total trainable parameters ($> 90\%$ of the total number of parameters).}. Most of the time the resulting algorithms are easy to execute on GPU-devices. 

\section{Preliminaries on circulant matrices}
\label{sec:circ}

In this paper, we use {\em circulant matrices} to build compact deep neural networks. A n-by-n circulant matrix $C$ is a matrix where each row is a cyclic right shift of the previous one as illustrated below.

\[
C = circ(c) =\left[\begin{array}{ccccc}
c_{0} & c_{n-1} & c_{n-2} & \dots & c_{1} \\
c_{1} & c_{0} & c_{n-1} & & c_{2} \\
c_{2} & c_{1} & c_{0}& & c_{3} \\
\vdots & & & \ddots & \vdots \\
c_{n-1} & c_{n-2} & c_{n-3} & & \phantom{0}c_{0}\phantom{0}
\end{array}\right]
\]

Because the circulant matrix $C \in \mathbb R^{n\times n}$ is fully determined by the vector $c \in \mathbb R^n$, the matrix $C$ can be compactly represented in memory using only $n$ real values instead of $n^2$.

An additional benefit of circulant matrices, is that they are computationally efficient, especially on GPU devices. Multiplying a circulant matrix $C$ by a vector $x$ is equivalent to a circular convolution between $c$ and $x$ (denoted $c \star x$). Furthermore, the circular convolution can be computed in the Fourier domain as follows. 

\begin{equation*}
  Cx \quad = \quad c \star x \quad = \quad \F^{-1}\left(\F(c) \times \F(x)\right)
\end{equation*}

\noindent where $\F$ is the Fourier transform. Because this operation can be simplified to a simple element wise vector multiplication, the matrix multiplication $Cx$ can be computed in $O(n \log n)$ instead of $O(n^2)$.

Among the many applications of circulant matrices, matrix decomposition is one of the interest. In particular, Schmid et al. have shown in \cite{muller1998algorithmic,schmid2000decomposing}, that any complex matrix $A \in \mathbb C^{n\times n}$ can be decomposed into the product of diagonal and circulant matrices, as follows: 

\begin{equation}\label{eq:general_framework}
  A = D^{(1)} C^{(1)} D^{(2)} C^{(2)} \dots D^{(m)} C^{(m)} = \prod_{i=1}^{m} D^{(i)} C^{(i)}
\end{equation}

Later in \cite{Huhtanen2015}, Huhtanen and Perämäki have demonstrated that choosing $m=n$ is sufficient to decompose  
any complex matrix $A \in \mathbb C^{n\times n}$. 
By~\cite{schmid2000decomposing}, the result in Equation~\ref{eq:general_framework} also holds for a real matrix $A \in \mathbb R^{n\times n}$, but the proof yields a much bigger value of $m$. However,  the construction of \cite{schmid2000decomposing} is far from optimal and it is likely that most real matrices can be decomposed into a reasonable number of factors. The authors of~\cite{moczulski2015acdc} made this conjecture, and they have leveraged the decomposition described in Equation~\ref{eq:general_framework} in order to implement compact fully connected layers.

\section{Compact  video classification architecture using circulant matrices}
\label{section:archi}

Building on the decomposition presented in the previous section and the previous results obtained in~\cite{moczulski2015acdc}, we now introduce a compact neural network architecture for video classification where dense matrices have been replaced by products of circulant and diagonal matrices.

\subsection{Base Model}

\begin{figure}[ht]
  \centering
  \tikzset{%
  >={Latex[width=2mm,length=2mm]},
            base/.style = {rectangle, draw=black, text centered, font=\sffamily},
             box/.style = {base, rounded corners, text depth=3cm, minimum height=4cm, minimum width=3cm},
     transparent/.style = {rectangle, draw=black},
       circulant/.style = {base, fill=yellow!30},
       embedding/.style = {base, fill=blue!30, minimum width=2.5cm, minimum height=1cm},
           other/.style = {base, fill=white!30,  minimum width=2cm, minimum height=1cm},
              fc/.style = {base, fill=orange!30, minimum width=1.5cm, minimum height=1cm},
          gating/.style = {base, fill=green!30, minimum width=2cm, text width=2cm, minimum height=1cm},
             moe/.style = {base, fill=purple!30, minimum width=1.5cm, minimum height=1cm},
}

\begin{tikzpicture}[every node/.style={fill=white, font=\sffamily}, align=center]

  \draw (0.0, +2.)  node [other, draw=none] {\textbf{Embedding}};
  \draw (+3.7, +2.)  node [other, draw=none] {\textbf{Dim Reduction}};
  \draw (+8.0, +2.)  node [other, draw=none] {\textbf{Classification}};

  \draw (0, +0.8)  node [embedding] {Video};
  \draw (0, -0.8)  node [embedding] {Audio};

  \draw (+2.5, +0.8)  node (fc) [fc] {FC};
  \draw (+2.5, -0.8)  node (fc) [fc] {FC};

  \draw (+4.75, 0)  node (fc) [other] {concat};
  \draw (+7.0, 0)  node (moe) [moe] {MoE};
  \draw (+9.25, 0)  node (gating2) [gating] {Context Gating};
 
  \draw (+1.5, +2) [dashed] -- (+1.5, -1.7);
  \draw (+6, +2) [dashed] -- (+6, -1.7);
  
\end{tikzpicture}
  \caption{This figure shows the architecture used for the experiences. The network samples at random video and audio frames from the input. The sample goes through an embedding layer and is reduced with a Fully Connected layer. The results are then concatenated and classified with a Mixture-of-Experts and Context Gating layer.}
  \label{fig:model_baseline}
\end{figure}
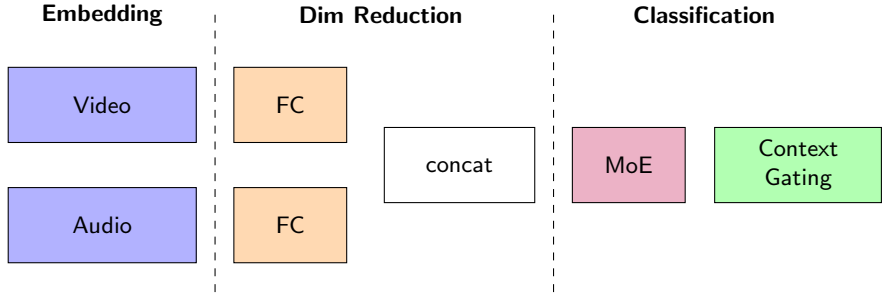

We demonstrate the benefit of circulant matrices using a base model which has been proposed by \cite{DBLP:journals/corr/MiechLS17}. This architecture can be decomposed into three blocks of layers, as illustrated in  Figure~\ref{fig:model_baseline}. 
The first block of layers, composed of the Deep Bag-of-Frames embedding, is meant to process audio and video frames independently.
The DBoF layer computes two embeddings: one for the audio and one for the video. In the next paragraph, we will only focus on describing the video embedding (The audio embedding is computed in a very similar way).

We represent a video $V$ as a set of $m$ frames $\{v_{1},\ldots,v_{m}\}$ where each frame $v_i \in \mathbb R^k$ is a vector of visual features extracted from the frame image.  In the context of the \yt competition, each $v_i$ is a vector of 1024 visual features extracted using the last fully connected layer of an Inception network trained on ImageNet.
The DBoF layer then embeds a video $V$ into a vector $v'$ drawn from a  $p$ dimensional vector space as follows:

\begin{equation*}
	v' = \max\left\{ Wv_{i}\mid v_i \in V \right\}
\end{equation*}

\noindent
where $W$ is a matrix in $\mathbb R^{p \times k}$ (learned) and max is the element-wise maximum operator. We typically choose $p > k$, (e.g. $p = 8192$).
Note that because this formulation is framed in term of sets, it can process videos of different lengths (i.e., a different value of $m$).

A second block of layers reduces the dimensionality of each embedding layer (audio and video), and merges the result into a single vector by using a simple concatenation operation. We chose to reduce the dimensionality of each embedding layer separately {\em before} the concatenation operation to avoid the concatenation of two high dimensional vectors.

Finally, the classification block uses a combination of Mixtures-of-Experts (MoE) and Context Gating to calculate the final probabilities. The Mixtures-of-Experts layer introduced in~\cite{716791} and proposed for video classification in~\cite{45619} is used to predict each label independently. It consists of a gating and experts networks which are concurrently learned. The gating network learns which experts to use for each label and the experts layers learn how to classify each label. The context gating operation was introduced in~\cite{DBLP:journals/corr/MiechLS17} and captures dependencies among features and re-weight probabilities based on the correlation of the labels. For example, it can capture the correlation of the labels {\em ski} and {\em snow} and re-adjust the probabilities accordingly. 

Table~\ref{table:shape_dbof} shows the shapes of the layers as well as the shapes of the weight matrices. 
\begin{table}[ht]
  \centering
  \begin{tabular}{L{2.5cm} | C{2cm} | C{2.5cm} | C{2.5cm} | C{2.5cm} }
    \toprule
    \multirow{2}{*}{\textbf{Layer}} & \multirow{2}{*}{\textbf{Layer Size}} & \textbf{Activation shape} & \textbf{Weight matrix shape} & \multirow{2}{*}{\textbf{\#Weights}} \\
    \midrule
    \midrule
    Video DBoF & 8192 & (-1, 150, 1024) & (1024, 8192) & \numprint{8388608} \\
	Audio DBoF & 4096 & (-1, 150, 128) & (128, 4096) & \numprint{524288} \\
    Video FC & 512 & (-1, 8192) & (8192, 512) & \numprint{4194304} \\
	Audio FC & 512 & (-1, 4096) & (4096, 512) & \numprint{2097152} \\
    Concat & - & (-1, 1024) & - & - \\
    MoE Gating & 3 & (-1, 1024) & (1024, 19310) & \numprint{19773440} \\
    MoE Experts & 2 & (-1, 1024) & (1024, 15448) & \numprint{15818752} \\
    Context Gating & - & (-1, 3862) & (3862, 3862) & \numprint{14915044} \\
   \bottomrule
  \end{tabular}
  \\[0.2cm]
  \setlength{\belowcaptionskip}{-0.2cm}
  \caption{This table shows the architecture of our base model with a DBoF Embedding and 150 frames sampled from the input. For more clarity, weights from batch normalization layers have been ignored. The $-1$ in the activation shapes corresponds to the batch size. The size of the MoE layers corresponds to the number of mixtures used.}
  \label{table:shape_dbof}
\end{table}

\subsection{Robust Deep Bag-of-Frames pooling method}
\label{subsection:robust_dbof}
We propose a technique to extract more performance from the base model with DBoF embedding. The maximum pooling is sensitive to outliers and noise whereas the average pooling is more robust. We propose a method which consists in taking several samples of frames, applying the upsampling followed by the maximum pooling to these samples, and then averaging over all samples.
More formally, assume a video contains $m$ frames $v_{1},\ldots,v_{m}\in\mathbb{R}^{1024}$.
We first draw $n$ random samples $S_{1}\ldots S_{n}$ of size $k$ 
from the set $\left\{ v_{1},\ldots,v_{m}\right\} $. The output
of the robust-DBoF layer is:
\begin{equation*}
  \frac{1}{n}\sum_{i=1}^{n}\max\left\{ v\times W:v\in S_{i}\right\} 
\end{equation*}
\noindent
Depending on $n$ and $k$, this pooling method is a tradeoff between the max pooling and the average pooling. Thus, it is more robust to noise, as will be shown in the experiments section.

\subsection{Compact representation of the base model}
\label{subsection:compact}

In order to train this model in a compact form we build upon the work of~\cite{7410684} and use a more general framework presented by  Equation~\ref{eq:general_framework}. The fully connected layers are then represented as follows:
$$h(x) = \phi\left(\left[\prod_{i=1}^{m} D^{(i)} C^{(i)}\right]x + b\right)$$
\noindent
where the parameters of each matrix $D^{(i)}$ and $C^{(i)}$ are trained using a gradient based optimization algorithm, and $m$ defines the number of factors. Increasing the value of $m$ increases the number of trainable parameters and therefore the modeling capabilities of the layer. In our experiments, we chose the number of factors $m$ empirically to achieve the best trade-off between model size and accuracy.

To measure the impact of the size of the model and its accuracy, we represent layers in their compact form independently. 
Given that circulant and diagonal matrices are square, we use concatenation and slicing to achieve the desired dimension. As such, with $m=1$, the weight matrix (1024, 8192) of the video embedding is represented by a concatenation of 8 DC matrices and the weight matrix of size (8192, 512) is represented by a single DC matrix with shape (8192, 8192) and the resulting output is sliced at the 512 dimension. We denote layers in their classic form as \textit{``Dense''} and layers represented with circulant and diagonal factors as \textit{``Compact''}.

\subsection{Leveraging architectural diversity}
\label{subsection:ensemble}

\begin{figure}[ht]
  \centering
  \scalebox{.75}{\tikzset{%
  >={Latex[width=2mm,length=2mm]},
            base/.style = {rectangle, draw=black, text centered, font=\sffamily},
             box/.style = {base, rounded corners, text depth=2cm, minimum height=2cm, minimum width=3cm},
     transparent/.style = {rectangle, draw=black},
        fc_small/.style = {base, fill=orange!30, minimum width=0.5cm},
       embedding/.style = {base, fill=blue!30, minimum width=2.5cm},
          concat/.style = {base, fill=white!30, minimum height=1cm},
           other/.style = {base, fill=white!30,  minimum width=1.7cm, minimum height=0.7cm},
              fc/.style = {base, fill=orange!30, minimum width=1.5cm, minimum height=1cm},
          gating/.style = {base, fill=green!30, minimum width=2cm, text width=2cm, minimum height=1cm},
             moe/.style = {base, fill=purple!30, minimum width=1.5cm, minimum height=1cm},
}
\begin{tikzpicture}[every node/.style={fill=white, font=\sffamily}, align=center]

  \draw (0,   0) node (box1) [box] {Video};
  \draw (0, +0.4)  node [embedding] {DBoF};
  \draw (0, -0.2)  node [embedding] {NetVLAD};
  \draw (0, -0.8)  node [embedding] {NetFV};
  \draw (+2.25, +0.4)  node (fc) [fc_small] {FC};
  \draw (+2.25, -0.2)  node (fc) [fc_small] {FC};
  \draw (+2.25, -0.8)  node (fc) [fc_small] {FC};
  
  \draw (0, -2.6) node (box1) [box] {Audio};
  \draw (0, +0.4-2.6)  node [embedding] {DBoF};
  \draw (0, -0.2-2.6)  node [embedding] {NetVLAD};
  \draw (0, -0.8-2.6)  node [embedding] {NetFV};
  \draw (+2.25, +0.4-2.6)  node (fc) [fc_small] {FC};
  \draw (+2.25, -0.2-2.6)  node (fc) [fc_small] {FC};
  \draw (+2.25, -0.8-2.6)  node (fc) [fc_small] {FC};
  
  \draw (+3.75, -0.2-2.6)  node (concat) [other] {average};
  \draw (+3.75, -0.2-0.0)  node (concat) [other] {average};
  \draw (+5.85, -0.8-0.7)  node (concat) [other] {concat};

  \draw (+7.95, -0.8-0.7)  node (moe) [moe] {MoE};
  \draw (+10.1, -0.8-0.7)  node (gating2) [gating] {Context Gating};
  
  \draw (+0, +1.75)  node [other, draw=none] {\textbf{Embedding}};
  \draw (+4.5, +1.75)  node [other, draw=none] {\textbf{Dim Reduction}};
  \draw (+9.2, +1.75)  node [other, draw=none] {\textbf{Classification}};
  
  \draw (+1.75, +2.0) [dashed] -- (+1.75, -4.0);
  \draw (+6.95, +2.0) [dashed] -- (+6.95, -4.0);
    
\end{tikzpicture}}
  \caption{This figure shows an evolution of the first architecture from figure~\ref{fig:model_baseline} with several embeddings. This architecture is made to leverage the diversity of an Ensemble in a single model.}
  \label{fig:diverstiy_architecture}
\end{figure}
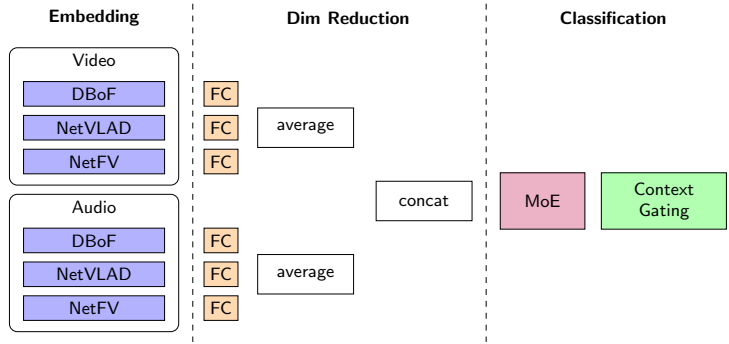

In order to benefit from architectural diversity, we also devise a single model architecture that combines different types of embedding layers. As we can see in Figure~\ref{fig:diverstiy_architecture}, video and audio frames are processed by several embedding layers before being reduced by a series of compact fully connected layers. The output of the compact fully connected layers are then averaged, concatenated and fed into the final classification block. Figure~\ref{fig:models} shows the result of different models given the number of parameters. The use of circulant matrices allow us to fit this model in GPU memory. For example, the diversity model with a NetVLAD embedding (cluster size of 256) and NetFV embedding (cluster size of 128) has 160 millions parameters (600 Mo) in the compact version and 728M (2.7 Go) in the dense version. 

\section{Experiments}
\label{sec:xp}

In this section, we first evaluate the pooling technique proposed in Section~\ref{subsection:robust_dbof}. Then, we conduct experiments to evaluate the accuracy of our compact models. In particular, we investigate which layer benefits the most from a circulant representation and show that the decomposition presented in Section~\ref{sec:circ} performs better than the  approach from~\cite{7410684} for the video classification problem. Finally, we compare all our models on a two dimensional size vs. accuracy scale in order to evaluate the trade-off between size and accuracy of each one of our models.

\subsection{Experimental Setup}
\paragraph*{\textbf{Dataset}}
All the experiments of this paper have been done in the context of the \textit{2nd YouTube-8M Video Understanding Challenge} with the \yt dataset. We trained our models with the full training set and 70\% of the validation set which corresponds to a total of \numprint{4822555} examples. We used the data augmentation technique proposed by~\cite{skalic2017deep} to virtually double the number of inputs. 
The method consists in splitting the videos into two equal parts. This approach is motivated by the observation  that a human could easily label the video by watching either the beginning or the ending of the video. 

All the code used in this experimental section is available online.\footnote{https://github.com/araujoalexandre/youtube8m-circulant}

\paragraph*{\textbf{Hyper-parameters}}
All our experiments are developed with TensorFlow Framework~\cite{tensorflow2015-whitepaper}. We trained our models with the CrossEntropy loss and used Adam optimizer with a 0.0002 learning rate and a 0.8 exponential decay every 4 million examples. All fully connected layers are composed of 512 units. DBoF, NetVLAD and NetFV are respectively 8192, 64 and 64 of cluster size for video frames and 4096, 32, 32 for audio frames. We used 4 mixtures for the MoE Layer. We used all the available 150 frames and robust max pooling introduced in~\ref{subsection:robust_dbof} for the DBoF embedding. In order to stabilize and accelerate the training, we used batch normalization before each non linear activation and gradient clipping. 

\paragraph*{\textbf{Evaluation Metric}}
We used the GAP (Global Average Precision), as used in the \textit{2nd YouTube-8M Video Understanding Challenge}, to compare our experiments. The GAP metric is defined as follows:
\begin{equation*}
	GAP = \sum_{i=1}^{P}p(i) \Delta r(i)
\end{equation*}
where $P$ is the number of final predictions, $p(i)$ the precision, and $r(i)$ the recall. We limit our evaluation to 20 predictions for each video. 

\paragraph*{\textbf{Hardware}}
All experiments have been realized on a cluster of 12 nodes. Each node has 160 POWER8 processor, 128 Go of RAM and 4 Nividia Titan P100.

\subsection{Robust Deep Bag-of-Frames pooling method}
\label{subsection:exp_bagging}
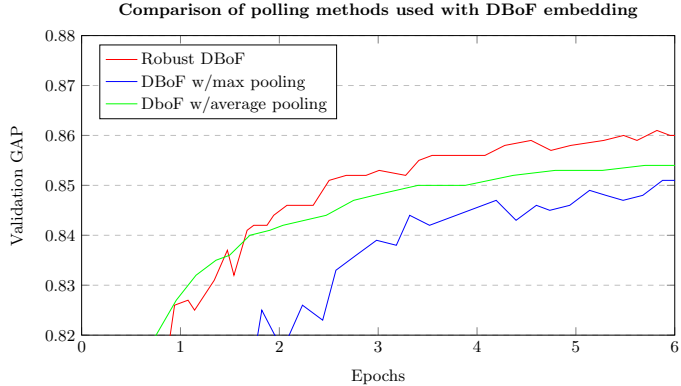
\begin{figure}[!htb]
  \centering
  \begin{tikzpicture}[scale=0.7]
\begin{axis}[
    width=0.85\textwidth,
    height=\axisdefaultheight,
    title={\textbf{Comparison of polling methods used with DBoF embedding}},
    legend style={fill=white,draw=black},
    legend cell align={left},
    xlabel={Epochs},
    ylabel={Validation GAP},
    xmin=0, xmax=6,
    ymin=0.82, ymax=0.88,
    xtick={0,1,2,3,4,5,6},
    ytick={0.82,0.83,0.84,0.85,0.86,0.87,0.88},
    legend pos=north west,
    ymajorgrids=true,
    grid style=dashed,
	]
  \addplot[color=red] table [y=gap, x=epoch]{robust_dbof/dbof_robust.dat};
  \addplot[color=blue] table [y=gap, x=epoch]{robust_dbof/dbof_max_pooling.dat};
  \addplot[color=green] table [y=gap, x=epoch]{robust_dbof/dbof_avg_pooling.dat};
\legend{
 Robust DBoF,
 DBoF w/max pooling, 
 DboF w/average pooling, 
 }
\end{axis}
\end{tikzpicture}
  \\[-0.3cm]
  \setlength{\belowcaptionskip}{-0.4cm}
  \caption{This graphic shows the impact of \textit{robust DBoF} (i.e. red line) with $n=10$ and $k=15$ on the Deep Bag-of-Frames embedding compared to max and average pooling.}
  \label{fig:learning_curve_bagging}
\end{figure}

We evaluate the performance of our Robust DBoF embedding. In accordance with the work from~\cite{45619}, we find that average pooling performs better than maximum pooling. 
Figure~\ref{fig:learning_curve_bagging} shows that the proposed robust maximum pooling method outperforms both maximum and average pooling.

\subsection{Impact of circulant matrices on different layers}
\begin{figure}[!htb]
  \centering
  \begin{tikzpicture}[scale=0.7]
\begin{axis}[
    width=0.85\textwidth,
    height=\axisdefaultheight,
    title={\parbox{8cm}{\centering \textbf{Comparison of the effect of compactness over different layers with the base model}}},
    legend columns=2,
    legend style={fill=white,
                  draw=black, 
    			 /tikz/column 2/.style={
                    column sep=5pt,
                  },},
    legend cell align={left},
    xlabel={Epochs},
    ylabel={Validation GAP},
    xmin=0, xmax=7,
    ymin=0.77, ymax=0.87,
    xtick={0,1,2,3,4,5,6,7},
    ytick={0.77,0.78,0.79,0.8,0.81,0.82,0.83,0.84,0.85,0.86,0.87},
    legend pos=north west,
    ymajorgrids=true,
    grid style=dashed,
	]
  \addplot[color=red] table [y=gap, x=epoch]{layers/dense.dat};
  \addplot[color=yellow] table [y=gap, x=epoch]{layers/compact_dbof.dat};
  \addplot[color=blue] table [y=gap, x=epoch]{layers/compact_fc.dat};
  \addplot[color=green] table [y=gap, x=epoch]{layers/compact_moe.dat};
\legend{
   Dense Model,
   Model with compact DBoF, 
   Model with compact FC, 
   Model with compact MoE,
 }
\end{axis}
\end{tikzpicture}
  \caption{Validation GAP according to the number of epochs for different compact models.}
  \label{fig:learning_curve_layers}
\end{figure}
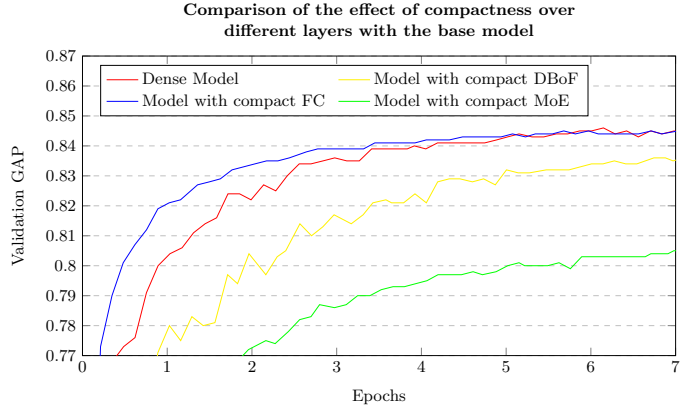

This series of experiments aims at understanding the effect of compactness over different layers. Table~\ref{table:circulant_layer} shows the result in terms of number of weights, size of the model (MB) and GAP. We also compute the compression ratio with respect to the dense model. The compact fully connected layer achieves a compression rate of 9.5 while having a very similar performance, whereas the compact DBoF and MoE achieve a higher compression rate at the expense of accuracy. 
Figure~\ref{fig:learning_curve_layers} shows that the model with a compact FC converges faster than the dense model. The model with a compact DBoF shows a big variance over the validation GAP which can be associated with a difficulty to train. The model with a compact MoE is more stable but at the expense of its performance.
Another series of experiments investigates the effect of adding factors of the compact matrix DC (i.e. the parameters $m$ specified in section~\ref{subsection:compact}). Table~\ref{table:factors} shows that there is no gain in accuracy even if the number of weights increases. It also shows that adding factors has an important effect on the  speed of training. On the basis of this result, i.e. given the performance and compression ratio, we can consider that representing the fully connected layer of the base model in a compact fashion can be a good trade-off.

\begin{table}[!htb]
  \centering
  \begin{tabular}{L{3cm} | C{2cm} | C{2cm} | C{2.5cm} | C{2cm} | C{2cm} }
    \toprule
    \multirow{2}{*}{\textbf{Baseline Model}} & \multirow{2}{*}{\textbf{\#Weights}} & \multirow{2}{*}{\textbf{Size (MB)}} & \textbf{Compress. Rate (\%)} & \multirow{2}{*}{\textbf{GAP@20}} & \multirow{2}{*}{\textbf{Diff.}} \\
    \midrule
    \midrule
	Dense Model & \numprint{45359764} & 173 & - & \textbf{0.846} & -\\
    Compact DBoF & \numprint{36987540} & 141 & 18.4 & 0.838 & -0.008\\
    Compact FC & \numprint{41181844} & 157 & 9.2 & 0.845 & \textbf{-0.001} \\
    Compact MoE & \numprint{12668504} & 48 & 72.0 & 0.805 & -0.041 \\
   \bottomrule
  \end{tabular}
  \\[0.2cm]
  \caption{This table shows the effect of the compactness of different layers. In these experiments, for speeding-up  the training phase, we did not use the audio features and exploited only the video information.}
  \label{table:circulant_layer}
\end{table}

\begin{table}[!htb]
  \centering
  \begin{tabular}{C{2cm} | C{2.6cm} | C{2.5cm} | C{2.8cm} | C{1.6cm}}
  \toprule
  \multirow{2}{*}{\textbf{\#factors}} & \multirow{2}{*}{\textbf{\#Examples/sec}} & \textbf{\#parameters in FC Layer} & \textbf{Compress. Rate of FC layer (\%)} & \multirow{2}{*}{\textbf{GAP@20}} \\
  \midrule
  \midrule
  1 & \numprint{1052} & \numprint{12288} & 99.8 & 0.861 \\
  3 & 858 & \numprint{73728} & 98.8 & 0.861 \\
  6 & 568 & \numprint{147456} & 97.6 & 0.859 \\
  Dense FC & \numprint{1007} & \numprint{6291456} & - & 0.861 \\
  \bottomrule
  \end{tabular}
  \\[0.2cm]
  \caption{This table shows the evolution of the number of parameters and the accuracy according to the number of factors. Despite the addition of degrees of freedom for the weight matrix of the fully connected layer, the model does not improve in performance. The column \textit{\#Examples/sec} shows the evolution of images per sec processed during the training of the model with a compact FC according to the number of factors.}
  \label{table:factors}
\end{table}

\subsection{Comparison with related works}
\begin{figure}[!htb]
  \centering
  \begin{tikzpicture}[scale=0.7]
\begin{axis}[
    width=0.85\textwidth,
    height=\axisdefaultheight,
    title={\textbf{GAP given the pooling method used with DBoF embedding}},
    legend style={fill=white,draw=black},
    legend cell align={left},
    xlabel={Epochs},
    ylabel={Validation GAP},
    xmin=0, xmax=10,
    ymin=0.79, ymax=0.88,
    xtick={0,1,2,3,4,5,6,7,8,9,10},
    ytick={0.79,0.80,0.81,0.82,0.83,0.84,0.85,0.86,0.87,0.88},
    legend pos=north west,
    ymajorgrids=true,
    grid style=dashed,
	]
  \addplot[color=red] table [y=gap, x=epoch]{dc_cd/dc.dat};
  \addplot[color=blue] table [y=gap, x=epoch]{dc_cd/cd.dat};
\legend{
 Compact FC w/general approach,
 {Compact FC w/CD and $D \in \{-1, 1\}$},
 }
\end{axis}
\end{tikzpicture}
  \caption{This figure shows the GAP difference between the $CD$ approach proposed in~\cite{7410684} and the more generalized $DC$ approach from section~\ref{subsection:compact}. Instead of having $D \in \{-1, +1\}$ fixed, the generalized approach allows $D$ to be learned.}
  \label{fig:learning_dc_cd}
\end{figure}
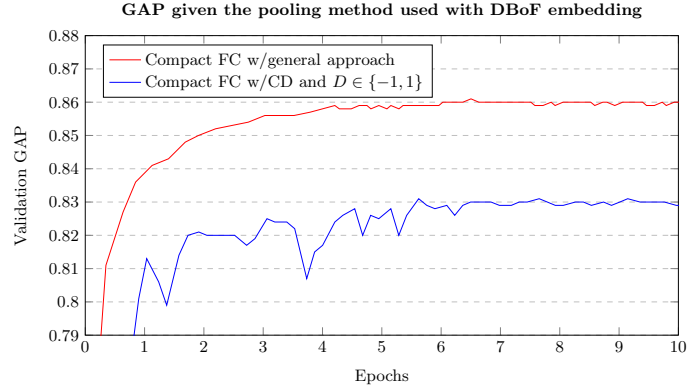

Circulant matrices have been used in neural networks in~\cite{7410684}. They proposed to replace fully connected layers by a circulant and diagonal matrices where the circulant matrix is learned by a gradient based optimization algorithm and the diagonal matrix is random with values in \{-1, 1\}. We compare our more general framework with their approach. Figure~\ref{fig:learning_dc_cd} shows the validation GAP according to the number of epochs of the base model with a compact fully connected layer implemented with both approaches.

\subsection{Compact Baseline model with different embeddings}
\begin{figure}[!htb]
  \centering
  \begin{tikzpicture}[scale=0.56]
\begin{axis}[
    title={\large \textbf{DBoF}},
    legend style={fill=none,draw=none},
    legend cell align={left},
    xlabel={Epochs},
    ylabel={Validation GAP},
    xmin=0, xmax=10,
    ymin=0.81, ymax=0.88,
    xtick={0,1,2,3,4,5,6,7,8,9,10},
    ytick={0.81, 0.82, 0.83,0.84,0.85,0.86, 0.87, 0.88},
    legend pos=north west,
    ymajorgrids=true,
    grid style=dashed,
	]
  \addplot[color=blue] table [y=gap, x=epoch]{fc_circulant_embedding/dbof_compressed.dat};
  \addplot[color=red] table [y=gap, x=epoch]{fc_circulant_embedding/dbof_uncompressed.dat};
\legend{Compact, Dense}
\end{axis}
\end{tikzpicture}
\begin{tikzpicture}[scale=0.56]
\begin{axis}[
    title={\large \textbf{NetVLAD}},
    legend style={fill=none,draw=none},
    legend cell align={left},
    xlabel={Epochs},
    ylabel={Validation GAP},
    xmin=0, xmax=10,
    ymin=0.80, ymax=0.88,
    xtick={0,1,2,3,4,5,6,7,8,9,10},
    ytick={0.81, 0.82, 0.83,0.84,0.85,0.86, 0.87, 0.88},
    legend pos=north west,
    ymajorgrids=true,
    grid style=dashed,
  ]
  \addplot[color=blue] table [y=gap, x=epoch]{fc_circulant_embedding/netvlad_compressed.dat};
  \addplot[color=red] table [y=gap, x=epoch]{fc_circulant_embedding/netvlad_uncompressed.dat};
\legend{Compact, Dense}
\end{axis}
\end{tikzpicture}
\begin{tikzpicture}[scale=0.56]
\begin{axis}[
    title={\large \textbf{NetFV}},
    legend style={fill=none,draw=none},
    legend cell align={left},
    xlabel={Epochs},
    ylabel={Validation GAP},
    xmin=0, xmax=10,
    ymin=0.810, ymax=0.88,
    xtick={0,1,2,3,4,5,6,7,8,9,10},
    ytick={0.81, 0.82, 0.83, 0.84, 0.85,0.86, 0.87, 0.88},
    legend pos=north west,
    ymajorgrids=true,
    grid style=dashed,
  ]
  \addplot[color=blue] table [y=gap, x=epoch]{fc_circulant_embedding/fisher_compressed.dat};
  \addplot[color=red] table [y=gap, x=epoch]{fc_circulant_embedding/fisher_uncompressed.dat};
\legend{Compact, Dense}
\end{axis}
\end{tikzpicture}
  \caption{The figures above show the validation GAP of \textmd{compact} and \textit{Dense} fully connected layer with different embeddings according to the number of epochs.}
  \label{fig:learning_curve_circulant}
\end{figure}
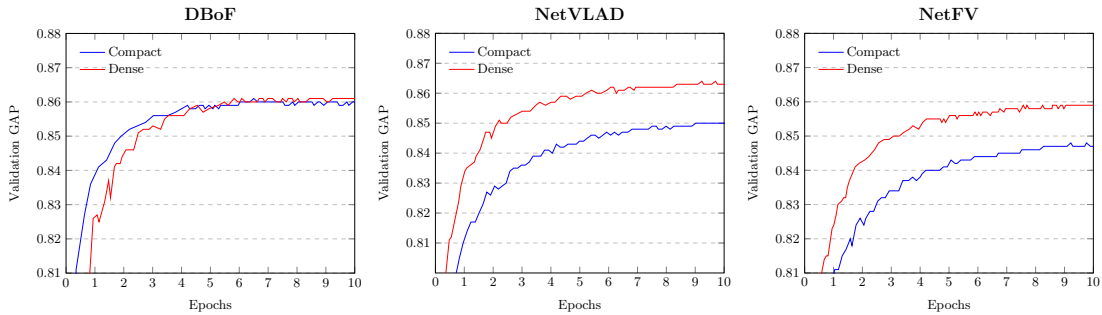

To compare the performance and the compression ratio we can expect, we consider different settings where the compact fully connected layer is used together with different embeddings. Figure~\ref{fig:learning_curve_circulant} and Table~\ref{table:fc_circulant_with_diff_embedding} show the performance of the base model with DBoF, NetVLAD and NetFV embeddings with a \textit{Dense} and \textit{Compact} fully connected layer. Notice that we can get a bigger compression rate with NetVLAD and NetFV due to the fact that the output of the embedding is in a higher dimensional space which implies a larger weight matrix for the fully connected layer. Although the compression rate is higher, it is at the expense of accuracy.

\begin{table}[!htb]
  \centering
  \begin{tabular}{L{3.5cm} C{3cm} C{2cm} C{2.5cm} C{2cm} }
    \toprule
    \multirow{2}{*}{\textbf{Method}} & \multirow{2}{*}{\textbf{\#Parameters}} & \multirow{2}{*}{\textbf{Size (MB)}} & \textbf{Compress. Rate (\%)} & \multirow{2}{*}{\textbf{GAP@20}} \\
    \midrule
    \textbf{DBoF} \\
    \midrule
	\multicolumn{1}{l|}{FC Dense} & \multicolumn{1}{c|}{\numprint{65795732}} & \multicolumn{1}{c|}{251} & \multicolumn{1}{c|}{-} & 0.861 \\
    \multicolumn{1}{l|}{FC Circulant} & \multicolumn{1}{c|}{\numprint{59528852}} & \multicolumn{1}{c|}{227} & \multicolumn{1}{c|}{9.56} & 0.861 \\
   \midrule
   \textbf{NetVLAD} \\
   \midrule
	\multicolumn{1}{l|}{FC Dense} & \multicolumn{1}{c|}{\numprint{86333460}} & \multicolumn{1}{c|}{330} & \multicolumn{1}{c|}{-} & 0.864 \\
    \multicolumn{1}{l|}{FC Circulant} & \multicolumn{1}{c|}{\numprint{50821140}} & \multicolumn{1}{c|}{194} & \multicolumn{1}{c|}{41.1} & 0.851 \\
   \midrule
   \textbf{NetFisher} \\
   \midrule
	\multicolumn{1}{l|}{FC Dense} & \multicolumn{1}{c|}{\numprint{122054676}} & \multicolumn{1}{c|}{466} & \multicolumn{1}{c|}{-} & 0.860 \\
    \multicolumn{1}{l|}{FC Circulant} & \multicolumn{1}{c|}{\numprint{51030036}} & \multicolumn{1}{c|}{195} & \multicolumn{1}{c|}{58.1} & 0.848 \\
   \bottomrule
  \end{tabular}
 \\[0.2cm]
  \caption{This table shows the impact of the compression of the fully connected layer of the model architecture shown in Figure~\ref{fig:model_baseline} with Audio and Video features vector and different types of embeddings. The variable compression rate is due to the different width of the output of the embedding.}
  \label{table:fc_circulant_with_diff_embedding}
\end{table}

\subsection{Model size vs. accuracy}

\begin{figure}[ht!]
  \centering
  \begin{tikzpicture}[scale=0.85]
  \begin{axis}[
    height=\axisdefaultheight,
    title={\textbf{Benchmark of compact models}},
    legend style={fill=white,draw=none},
    legend cell align={left},
    xlabel={Number of weights (Millions)},
    ylabel={Validation GAP},
    xmin=0, xmax=200,
    ymin=0.81, ymax=0.88,
    xtick={0, 20, 40,60,80,100,120,140,160,180,200},
    ytick={0.81,0.82,0.83,0.84,0.85,0.86,0.87,0.88},
    legend pos=outer north east,
    ymajorgrids=true,
    grid style=dashed,
  ]
    \addplot[red,    only marks] coordinates {( 25, 0.831)};
    \addplot[orange, only marks] coordinates {( 50, 0.851)};
    \addplot[brown,  only marks] coordinates {( 51, 0.848)};
    \addplot[green,  only marks] coordinates {( 59, 0.861)};
    \addplot[lime,   only marks] coordinates {( 87, 0.858)};
    \addplot[olive,  only marks] coordinates {(158, 0.861)};
    \addplot[blue,   only marks] coordinates {(159, 0.863)};
    \addplot[cyan,   only marks] coordinates {(166, 0.861)};
    \addplot[teal,   only marks] coordinates {(176, 0.861)};
    \legend{
      {NetVLAD 32, Compact FC 256, MoE 2},
      {NetVLAD 64, Compact FC 512, MoE 4},
      {NetFV 64, Compact FC 512, MoE 4},
      {DBoF 8192, Compact FC 512, MoE 4},
      {NetVLAD 256, Compact FC 1024, MoE 4},
      {NetVLAD 128, NetFV 64, Compact FC 1024, MoE 4},
      {NetVLAD 256, NetFV 128, Compact FC 1024, MoE 4},
      {DBoF 8192,  NetVLAD 128, Compact FC 1024, MoE 4},
      {DBoF 16384, NetVLAD 256, Compact FC 1024, MoE 4},
    }
  \end{axis}
\end{tikzpicture}
  \caption{Comparison between different models with compact fully connected layers.}
  \label{fig:models}
\end{figure}
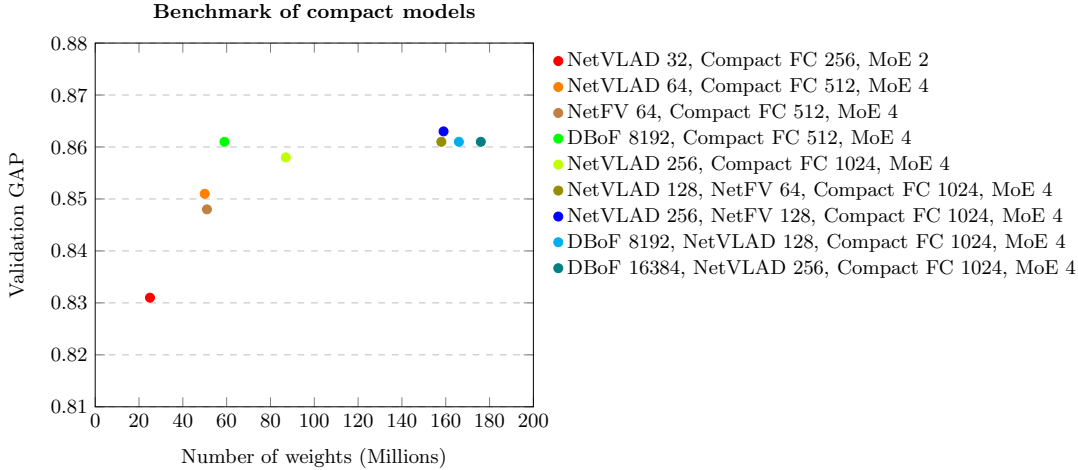

To conclude our experimental evaluation, we compare all our models in terms of size and accuracy. The results are presented in Figure~\ref{fig:models}. 

As we can see in this figure, the most compact models are obtained with the circulant NetVLAD and NetFV. We can also see that the complex architectures described in Section~\ref{subsection:ensemble} (DBoF + NetVLAD) achieve top performance but at the cost of a large number of weights. Finally, the best trade-off between size and accuracy is obtained using the DBoF embedding layer and achieves a GAP of 0.861 for only 60 millions weights. 

\section{Conclusion}
In this paper, we demonstrated that circulant matrices can be a great tool to design compact neural network architectures for video classification tasks. We proposed a more general framework which improves the state of the art and conducted a series of experiments aiming at understanding the effect of compactness on different layers. Our experiments demonstrate that the best trade-off between size and accuracy is obtained using circulant DBoF embedding layers. We investigated a model with multiple embeddings to leverage the performance of an Ensemble but found it ineffective.
The good performance of Ensemble models, i.e. why aggregating different distinct models performs better that incorporating all the diversity in a single architecture  is still an open problem. Our future work will be devoted to address this challenging question and to pursue our effort to devise compact models achieving the same accuracy as larger one, and to study their theoretical properties.

\section{Acknowledgement}
This work was granted access to the OpenPOWER prototype from GENCI-IDRIS under the Preparatory Access AP010610510 made by GENCI. We would like to thank the staff of IDRIS who was really available for the duration of this work, Abdelmalek Lamine and Tahar Nguira, interns at Wavestone for their work on circulant matrices. Finally, we would also like to thank Wavestone to support this research. 


%
\bibliographystyle{splncs04}
\bibliography{mybibliography}

\begin{thebibliography}{10}
\providecommand{\url}[1]{\texttt{#1}}
\providecommand{\urlprefix}{URL }
\providecommand{\doi}[1]{https://doi.org/#1}

\bibitem{tensorflow2015-whitepaper}
Abadi, M., Agarwal, A., Barham, P., Brevdo, E., Chen, Z., Citro, C., Corrado,
  G.S., Davis, A., Dean, J., Devin, M., Ghemawat, S., Goodfellow, I., Harp, A.,
  Irving, G., Isard, M., Jia, Y., Jozefowicz, R., Kaiser, L., Kudlur, M.,
  Levenberg, J., Man\'{e}, D., Monga, R., Moore, S., Murray, D., Olah, C.,
  Schuster, M., Shlens, J., Steiner, B., Sutskever, I., Talwar, K., Tucker, P.,
  Vanhoucke, V., Vasudevan, V., Vi\'{e}gas, F., Vinyals, O., Warden, P.,
  Wattenberg, M., Wicke, M., Yu, Y., Zheng, X.: {TensorFlow}: Large-scale
  machine learning on heterogeneous systems (2015),
  \url{https://www.tensorflow.org/}, software available from tensorflow.org

\bibitem{45619}
Abu-El-Haija, S., Kothari, N., Lee, J., Natsev, A.P., Toderici, G.,
  Varadarajan, B., Vijayanarasimhan, S.: Youtube-8m: A large-scale video
  classification benchmark. In: arXiv:1609.08675 (2016),
  \url{https://arxiv.org/pdf/1609.08675v1.pdf}

\bibitem{Arandjelovic16}
Arandjelovi\'c, R., Gronat, P., Torii, A., Pajdla, T., Sivic, J.: {NetVLAD}:
  {CNN} architecture for weakly supervised place recognition. In: IEEE
  Conference on Computer Vision and Pattern Recognition (2016)

\bibitem{bucilua2006model}
Buciluǎ, C., Caruana, R., Niculescu-Mizil, A.: Model compression. In:
  Proceedings of the 12th ACM SIGKDD international conference on Knowledge
  discovery and data mining. pp. 535--541. ACM (2006)

\bibitem{Chen:2015:CNN:3045118.3045361}
Chen, W., Wilson, J.T., Tyree, S., Weinberger, K.Q., Chen, Y.: Compressing
  neural networks with the hashing trick. In: Proceedings of the 32Nd
  International Conference on International Conference on Machine Learning -
  Volume 37. pp. 2285--2294. ICML'15, JMLR.org (2015),
  \url{http://dl.acm.org/citation.cfm?id=3045118.3045361}

\bibitem{7410684}
Cheng, Y., Yu, F.X., Feris, R.S., Kumar, S., Choudhary, A., Chang, S.F.: An
  exploration of parameter redundancy in deep networks with circulant
  projections. In: 2015 IEEE International Conference on Computer Vision
  (ICCV). pp. 2857--2865 (Dec 2015)

\bibitem{Collins2014MemoryBD}
Collins, M.D., Kohli, P.: Memory bounded deep convolutional networks. CoRR
  \textbf{abs/1412.1442} (2014)

\bibitem{Courbariaux:2015:BTD:2969442.2969588}
Courbariaux, M., Bengio, Y., David, J.P.: Binaryconnect: Training deep neural
  networks with binary weights during propagations. In: Proceedings of the 28th
  International Conference on Neural Information Processing Systems - Volume 2.
  pp. 3123--3131. NIPS'15, MIT Press, Cambridge, MA, USA (2015),
  \url{http://dl.acm.org/citation.cfm?id=2969442.2969588}

\bibitem{pmlr-v80-dai18d}
Dai, B., Zhu, C., Guo, B., Wipf, D.: Compressing neural networks using the
  variational information bottleneck. In: Dy, J., Krause, A. (eds.) Proceedings
  of the 35th International Conference on Machine Learning. Proceedings of
  Machine Learning Research, vol.~80, pp. 1143--1152. PMLR, Stockholmsmässan,
  Stockholm Sweden (10--15 Jul 2018),
  \url{http://proceedings.mlr.press/v80/dai18d.html}

\bibitem{NIPS2013_5025}
Denil, M., Shakibi, B., Dinh, L., Ranzato, M.A., de~Freitas, N.: Predicting
  parameters in deep learning. In: Burges, C.J.C., Bottou, L., Welling, M.,
  Ghahramani, Z., Weinberger, K.Q. (eds.) Advances in Neural Information
  Processing Systems 26, pp. 2148--2156. Curran Associates, Inc. (2013),
  \url{http://papers.nips.cc/paper/5025-predicting-parameters-in-deep-learning.pdf}

\bibitem{Gupta:2015:DLL:3045118.3045303}
Gupta, S., Agrawal, A., Gopalakrishnan, K., Narayanan, P.: Deep learning with
  limited numerical precision. In: Proceedings of the 32Nd International
  Conference on International Conference on Machine Learning - Volume 37. pp.
  1737--1746. ICML'15, JMLR.org (2015),
  \url{http://dl.acm.org/citation.cfm?id=3045118.3045303}

\bibitem{han2015deep_compression}
Han, S., Mao, H., Dally, W.J.: Deep compression: Compressing deep neural
  networks with pruning, trained quantization and huffman coding. International
  Conference on Learning Representations (ICLR)  (2016)

\bibitem{44873}
Hinton, G., Vinyals, O., Dean, J.: Distilling the knowledge in a neural
  network. In: NIPS Deep Learning and Representation Learning Workshop (2015),
  \url{http://arxiv.org/abs/1503.02531}

\bibitem{Huhtanen2015}
Huhtanen, M., Per{\"a}m{\"a}ki, A.: Factoring matrices into the product of
  circulant and diagonal matrices. Journal of Fourier Analysis and Applications
   \textbf{21}(5),  1018--1033 (Oct 2015). \doi{10.1007/s00041-015-9395-0},
  \url{https://doi.org/10.1007/s00041-015-9395-0}

\bibitem{Jaderberg2014SpeedingUC}
Jaderberg, M., Vedaldi, A., Zisserman, A.: Speeding up convolutional neural
  networks with low rank expansions. CoRR  \textbf{abs/1405.3866} (2014)

\bibitem{jegou:inria-00548637}
J{\'e}gou, H., Douze, M., Schmid, C., P{\'e}rez, P.: {Aggregating local
  descriptors into a compact image representation}. In: {CVPR 2010 - 23rd IEEE
  Conference on Computer Vision \& Pattern Recognition}. pp. 3304--3311. {IEEE
  Computer Society}, San Francisco, United States (Jun 2010).
  \doi{10.1109/CVPR.2010.5540039}, \url{https://hal.inria.fr/inria-00548637}

\bibitem{FCVID}
Jiang, Y.G., Wu, Z., Wang, J., Xue, X., Chang, S.F.: Exploiting feature and
  class relationships in video categorization with regularized deep neural
  networks. {IEEE} Transactions on Pattern Analysis and Machine Intelligence
  \textbf{40}(2),  352--364 (2018). \doi{10.1109/TPAMI.2017.2670560},
  \url{https://doi.org/10.1109/TPAMI.2017.2670560}

\bibitem{716791}
Jordan, M.I., Jacobs, R.A.: Hierarchical mixtures of experts and the em
  algorithm. In: Proceedings of 1993 International Conference on Neural
  Networks (IJCNN-93-Nagoya, Japan). vol.~2, pp. 1339--1344 vol.2 (Oct 1993).
  \doi{10.1109/IJCNN.1993.716791}

\bibitem{karpathy2014large}
Karpathy, A., Toderici, G., Shetty, S., Leung, T., Sukthankar, R., Fei-Fei, L.:
  Large-scale video classification with convolutional neural networks. In:
  Proceedings of the IEEE conference on Computer Vision and Pattern
  Recognition. pp. 1725--1732 (2014)

\bibitem{DBLP:journals/corr/LiGLBLLLZW17}
Li, F., Gan, C., Liu, X., Bian, Y., Long, X., Li, Y., Li, Z., Zhou, J., Wen,
  S.: Temporal modeling approaches for large-scale youtube-8m video
  understanding. CoRR  \textbf{abs/1707.04555} (2017)

\bibitem{NIPS2017_6813}
Lin, J., Rao, Y., Lu, J., Zhou, J.: Runtime neural pruning. In: Guyon, I.,
  Luxburg, U.V., Bengio, S., Wallach, H., Fergus, R., Vishwanathan, S.,
  Garnett, R. (eds.) Advances in Neural Information Processing Systems 30, pp.
  2181--2191. Curran Associates, Inc. (2017),
  \url{http://papers.nips.cc/paper/6813-runtime-neural-pruning.pdf}

\bibitem{7298681}
Liu, B., Wang, M., Foroosh, H., Tappen, M., Penksy, M.: Sparse convolutional
  neural networks. In: 2015 IEEE Conference on Computer Vision and Pattern
  Recognition (CVPR). pp. 806--814 (June 2015). \doi{10.1109/CVPR.2015.7298681}

\bibitem{DBLP:journals/corr/MellempudiKM0KD17}
Mellempudi, N., Kundu, A., Mudigere, D., Das, D., Kaul, B., Dubey, P.: Ternary
  neural networks with fine-grained quantization. CoRR  \textbf{abs/1705.01462}
  (2017)

\bibitem{DBLP:journals/corr/MiechLS17}
Miech, A., Laptev, I., Sivic, J.: Learnable pooling with context gating for
  video classification. CoRR  \textbf{abs/1706.06905} (2017)

\bibitem{moczulski2015acdc}
Moczulski, M., Denil, M., Appleyard, J., de~Freitas, N.: Acdc: A structured
  efficient linear layer. arXiv preprint arXiv:1511.05946  (2015)

\bibitem{muller1998algorithmic}
M{\"u}ller-Quade, J., Aagedal, H., Beth, T., Schmid, M.: Algorithmic design of
  diffractive optical systems for information processing. Physica D: Nonlinear
  Phenomena  \textbf{120}(1-2),  196--205 (1998)

\bibitem{4270291}
Perronnin, F., Dance, C.: Fisher kernels on visual vocabularies for image
  categorization. In: 2007 IEEE Conference on Computer Vision and Pattern
  Recognition. pp.~1--8 (June 2007). \doi{10.1109/CVPR.2007.383266}

\bibitem{rastegariECCV16}
Rastegari, M., Ordonez, V., Redmon, J., Farhadi, A.: Xnor-net: Imagenet
  classification using binary convolutional neural networks. In: ECCV (2016)

\bibitem{schmid2000decomposing}
Schmid, M., Steinwandt, R., M{\"u}ller-Quade, J., R{\"o}tteler, M., Beth, T.:
  Decomposing a matrix into circulant and diagonal factors. Linear Algebra and
  its Applications  \textbf{306}(1-3),  131--143 (2000)

\bibitem{NIPS2015_5869}
Sindhwani, V., Sainath, T., Kumar, S.: Structured transforms for
  small-footprint deep learning. In: Cortes, C., Lawrence, N.D., Lee, D.D.,
  Sugiyama, M., Garnett, R. (eds.) Advances in Neural Information Processing
  Systems 28, pp. 3088--3096. Curran Associates, Inc. (2015),
  \url{http://papers.nips.cc/paper/5869-structured-transforms-for-small-footprint-deep-learning.pdf}

\bibitem{skalic2017deep}
Skalic, M., Pekalski, M., Pan, X.E.: Deep learning methods for efficient large
  scale video labeling. arXiv preprint arXiv:1706.04572  (2017)

\bibitem{VYBIRAL20111096}
Vybíral, J.: A variant of the johnson–lindenstrauss lemma for circulant
  matrices. Journal of Functional Analysis  \textbf{260}(4),  1096 -- 1105
  (2011). \doi{https://doi.org/10.1016/j.jfa.2010.11.014},
  \url{http://www.sciencedirect.com/science/article/pii/S0022123610004507}

\bibitem{DBLP:journals/corr/WangZW17}
Wang, H., Zhang, T., Wu, J.: The monkeytyping solution to the youtube-8m video
  understanding challenge. CoRR  \textbf{abs/1706.05150} (2017)

\bibitem{7410530}
Yang, Z., Moczulski, M., Denil, M., d.~Freitas, N., Smola, A., Song, L., Wang,
  Z.: Deep fried convnets. In: 2015 IEEE International Conference on Computer
  Vision (ICCV). pp. 1476--1483 (Dec 2015). \doi{10.1109/ICCV.2015.173}

\bibitem{8099498}
Yu, X., Liu, T., Wang, X., Tao, D.: On compressing deep models by low rank and
  sparse decomposition. In: 2017 IEEE Conference on Computer Vision and Pattern
  Recognition (CVPR). pp. 67--76 (July 2017). \doi{10.1109/CVPR.2017.15}

\bibitem{yue2015beyond}
Yue-Hei~Ng, J., Hausknecht, M., Vijayanarasimhan, S., Vinyals, O., Monga, R.,
  Toderici, G.: Beyond short snippets: Deep networks for video classification.
  In: Proceedings of the IEEE conference on computer vision and pattern
  recognition. pp. 4694--4702 (2015)

\end{thebibliography}

\end{document}